\documentclass{article}

\usepackage{arxiv}

\usepackage[utf8]{inputenc} 
\usepackage[T1]{fontenc}    
\usepackage{hyperref}       
\usepackage{url}            
\usepackage{booktabs}       
\usepackage{amsfonts}       
\usepackage{nicefrac}       
\usepackage{microtype}      
\usepackage{lipsum}

\usepackage{authblk}

\usepackage{cite}
\usepackage[pdftex]{graphicx}
	 \usepackage{epstopdf}
   \DeclareGraphicsExtensions{.pdf,.jpg,.jpeg,.png,-eps-converted-to.pdf}
\usepackage[abs]{overpic}
\usepackage{amsfonts}
\usepackage{tikz}
\usetikzlibrary{arrows,shapes,positioning,fit,patterns,decorations.pathmorphing}
\usepackage[export]{adjustbox}
\usepackage{array}
\usepackage[caption=false,font=normalsize,labelfont=sf,textfont=sf,subrefformat=parens,labelformat=parens]{subfig}
\usepackage{etex}

\usepackage{threeparttable}
\usepackage{booktabs}
\usepackage{multirow}
\usepackage{tabularx}

\usepackage{tablefootnote}

\usepackage{amsmath,amssymb}
\usepackage{color}
\usepackage{xspace}
\usepackage[normalem]{ulem}
\usepackage{bm}

\makeatletter
\DeclareRobustCommand\onedot{\futurelet\@let@token\@onedot}
\def\@onedot{\ifx\@let@token.\else.\null\fi\xspace}
\def\etal{~et~al\onedot}
 
\def\ie{i.e\onedot}

\makeatother

\def\clap#1{\hbox to 0pt{\hss #1\hss}}%
\def\initials#1{\protect\clap{\smash{\raisebox{1.4ex}{\tiny{\textsf{\textit{#1}}}}}}}%
\makeatletter
\newcommand{\EDIT}[4][]{\protect\@ifundefined{hidecomments}{%
  \strut{\color{#3}{\hspace{0pt}\initials{#2}\protect\sout{#1}{~#4}}}%
  }{#4}}
\newcommand{\NOTEboxed}[3]{\protect\@ifundefined{hidecomments}{%
  {\begin{center}\fbox{\parbox{0.97\linewidth}{\protect\EDIT{#1}{#2}{#3}}}\end{center}}
  }{}}
\newcommand{\COMM}[3]{\protect\@ifundefined{hidecomments}{%
  {\protect\EDIT{#1}{#2}{#3}}
  }{}}
\newcommand{\DefAuthor}[2] 
{%
  \expandafter\newcommand\csname #1edit\endcsname[2][]{\protect\EDIT[##1]{#1}{#2}{##2}}
  \expandafter\newcommand\csname #1\endcsname[1]{\protect\COMM{#1}{#2}{[##1]}}
  \expandafter\newcommand\csname #1boxed\endcsname[1]{\NOTEboxed{#1}{#2}{##1}}
}
\definecolor{dfltgreen}       {rgb}{0.0,0.5,0.0}
\definecolor{dfltred}         {rgb}{0.7,0.0,0.0}
\newcommand{\REVadd}[1]{\protect\@ifundefined{hidecomments}{%
  \strut{\color{dfltgreen}{#1}}}{#1}}
\newcommand{\REVedit}[2][]{\protect\@ifundefined{hidecomments}{%
  \strut{\color{dfltred}{\protect\sout{#1}}\color{dfltgreen}{~#2}}}%
  {#2}}
\makeatother

\DeclareMathOperator*{\sinc}{sinc}

\newcommand{\fun}{f}
\newcommand{\data}{g}

\newcommand{\net}{\mathcal{G}}
\newcommand{\tm}{\textsuperscript{\textregistered} }
\newcommand{\param}{\vec{\overline{p}}}
\newcommand{\paramt}{\vec{P}}
\newcommand{\pos}{\vec{x}}

\title{Training Auto-Encoder-Based Optimizers for Terahertz Image Reconstruction}

\author[1, 2]{Tak~Ming~Wong}
\author[1, 3]{Matthias~Kahl}
\author[1, 3]{Peter~Haring~Bol\'ivar}
\author[1, 2]{Andreas~Kolb}
\author[1, 4]{Michael~M\"oller}

\affil[1]{Center for Sensor Systems (\emph{ZESS}), University of Siegen, 57076 Siegen, Germany}
\affil[2]{Computer Graphics and Multimedia Systems Group, University of Siegen, 57076 Siegen, Germany}
\affil[3]{Institute for High Frequency and Quantum Electronics (\emph{HQE}), University of Siegen, 57068 Siegen, Germany}
\affil[4]{Computer Vision Group, University of Siegen, 57076 Siegen, Germany}

\date{July 2, 2019}

\begin{document}
\maketitle


\begin{abstract}
 Terahertz (THz) sensing is a promising imaging technology for a wide variety of different applications. Extracting the interpretable and physically meaningful parameters for such applications, however, requires solving an inverse problem in which a model function determined by these parameters needs to be fitted to the measured data. Since the underlying optimization problem is nonconvex and very costly to solve, we propose learning the prediction of suitable parameters from the measured data directly. More precisely, we develop a model-based autoencoder in which the encoder network predicts suitable parameters and the decoder is fixed to a physically meaningful model function, such that we can train the encoding network in an unsupervised way. We illustrate numerically that the resulting network is more than 140 times faster than classical optimization techniques while making predictions with only slightly higher objective values. Using such predictions as starting points of local optimization techniques allows us to converge to better local minima about twice as fast as optimizing without the network-based initialization. 
\end{abstract}


\section{Introduction}
Terahertz (THz) imaging is an emerging sensing technology with a great potential for hidden object imaging, contact-free analysis, non-destructive testing and stand-off detection in various application fields, including semi-conductor industry, biological and medical analysis, material and quality control, safety and security~\cite{chan2007imaging,Jansen2010,siegel2002terahertz}. The physically interpretable quantities relevant to the aforementioned applications, however, cannot always be measured directly. Instead, in THz imaging systems, each pixel contains implicit information about such quantities, making the \textit{inverse problem} of inferring these physical quantities a challenging problem with high practical relevance.

As we will discuss in Sec.~\ref{sec:THz}, at each pixel location $\pos$ the relation between the desired (unknown) parameters $\param(\pos)=(\hat{e}(\pos),\sigma(\pos), \mu(\pos),\phi(\pos))\in \mathbb{R}^4$, i.e., the electric field amplitude $\hat{e}$, the position of the surface $\mu$, the width of the reflected pulse $\sigma$, and the phase $\phi$, and the actual measurements $\data(\pos)\in \mathbb{R}^{n_z}$ can be modelled via the equation $\data(\pos,z) = (\fun_{\hat{e},\sigma,\mu,\phi}(z_i))_{i \in \{1,\hdots,n_z\}} + \text{noise}$, where
\begin{eqnarray}
\label{eq:thzEquation}
&  & \fun_{\hat{e},\sigma,\mu,\phi}(z) = \hat{e} \sinc\left(\sigma (z-\mu)\right) \exp\left(-i(\omega z - \phi)\right), \\
  & & \sinc(t)  = \begin{cases} \cfrac{\sin(\pi t)}{\pi t} & t \neq 0, \\
    1 & t = 0,
  \end{cases}
\end{eqnarray}
and  $(z_i)_{i \in \{1,\hdots,n_z\}}$ is a device-dependent sampling grid $z_{grid}$.
More details of the THz model are described in~\cite{wong2019computational}. Thus, the crucial step in THz imaging is the solution of optimization problem of the form
\begin{equation}
\label{eq:thzOptimization}
 \min_{\hat{e},\sigma,\mu,\phi} \quad \text{Loss}(\fun_{\hat{e},\sigma,\mu,\phi}(z_{grid}), g(\pos)),
\end{equation}
at each pixel $\pos$, possibly along with additional regularizers on the unknown parameters. 
Even with simple choices of the loss function such as an $\ell^2$-squared loss, the resulting fitting problem is highly nonconvex and global solutions become rather expensive. Considering that the number $(n_x\cdot n_y)$ of pixels, i.e., of optimization problem \eqref{eq:thzOptimization} to be solved, typically is in the order of hundred thousands to millions, even local first order or quasi-Newton methods become quite costly: For example, running the build-in Trust-Region solver of MATLAB\tm to reconstruct a $446 \times 446$ THz image takes over 170 minutes. 

In this paper, we propose to train a neural network to solve the per-pixel optimization problem \eqref{eq:thzOptimization} directly. We formulate the training of the network as a model-based autoencoder (AE), which allows us to train the corresponding network with real data in an unsupervised way, i.e., without ground truth. We demonstrate that the resulting optimization network yields parameters $(\hat{e},\sigma,\mu,\phi)$ that result in only slightly higher losses than actually running an optimization algorithm, despite the advantage of being more than 140 times faster. Moreover, we demonstrate that our network can serve as an excellent initialization scheme for classical optimizers. By using the network's prediction as a starting point for a gradient-based optimizer, we obtain lower losses and converge more than 2x faster than classical optimization approaches, while benefiting from all theoretical guarantees of the respective minimization algorithm.  

This paper is organized as follows: Sec.~\ref{sec:THz} gives more details on how THz imaging systems work. Sec.~\ref{sec:related} summarizes the related work on learning optimizers, machine learning for THz imaging techniques, and model-based autoencoders. Sec.~\ref{sec:ourArchi} describes model-based AEs in contrast to classical supervised learning approaches in detail, before Sec.~\ref{sec:implementation} summarizes our implementation. Sec.~\ref{sec:experiments} compares the proposed approaches to classical (optimization-based) reconstruction techniques in terms of speed and accuracy before Sec.~\ref{sec:conclusions} draws conclusions.

\section{THz Imaging Systems}
\label{sec:THz}
There are several approaches to realizing THz imaging, e.g. femtosecond laser based scanning system~\cite{cooper2011thz,hu1995imaging}, synthetic aperture systems~\cite{mcclatchey2001time,ding2013thz}, and hybrid systems \cite{kahl2012}. 
A typical approach to THz imaging is based on the Frequency Modulated Continuous Wave (FMCW) concept~\cite{ding2013thz}, which uses active frequency modulated THz signals to sense reflected signals from the object. The reflected energy and phase shifts due to the signal path length make 3D THz imaging possible.

\begin{figure}[thb]
    \centering
    \includegraphics[width=0.7\textwidth]{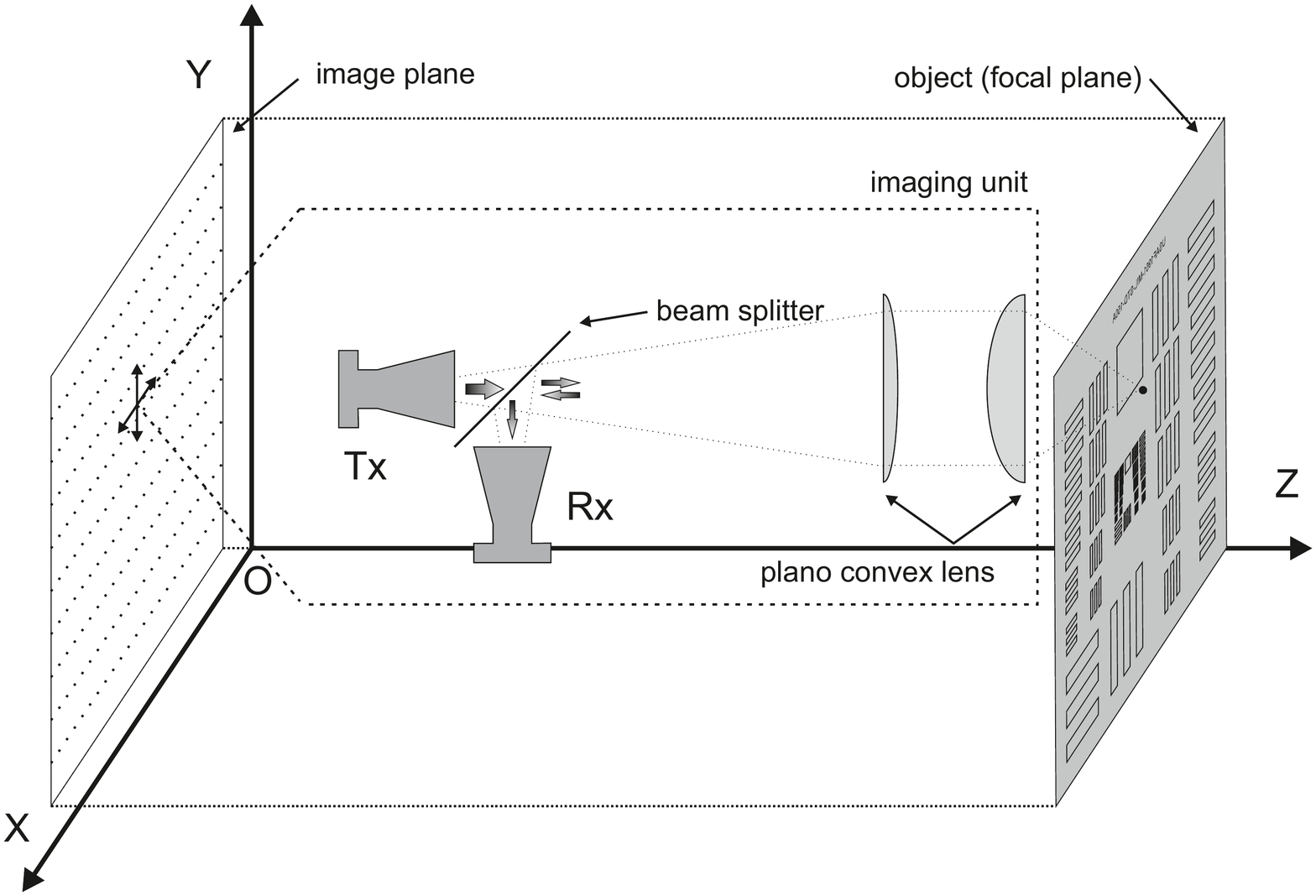}
    \caption{THz 3D imaging geometry. Both transmitter (Tx) and receiver (Rx) are mounted on the same platform. The imaging unit, consisting of Tx, Rx and optical components, are moved along the x and y direction using stepper motors and linear stages. This imaging unit takes a depth profile of the object at each lateral position, in order to acquire a full THz 3D image.}
    \label{fig:thz_geometry}
\end{figure}

In Figure \ref{fig:thz_geometry}, the setup of our electronic FMCW-THz 3D imaging system is shown. More details on the THz imaging system are described in \cite{ding2013thz}.

In this paper, we denote by $\data_t(\pos,t)$ the measured demodulated time domain signal of the reflected electric field amplitude of the FMCW system at lateral position $\pos\in \mathbb{R}^2$. 
In FMCW radar signal processing, this continuous wave temporal signal is converted into frequency domain by a Fourier transform~\cite{munson1989signal,skolnik1970radar}.
Since the linear frequency sweep has a unique frequency at each spatial position in $z$-direction, the converted frequency domain signal directly relates to the spatial azimuth ($z$-direction) domain signal
\begin{equation}
\label{eq:thzEquation_fft}
    \data_c(\pos,z) = \mathcal{F}\{\data_{t}(\pos,t)\}.
\end{equation}
The resulting 3D image $\data_c \in \mathbb{C}^{n_x \times n_y \times n_z}$ is complex data in the spatial domain, representing per-pixel complex reflectivity of THz energy.  The quantities $n_x$, $n_y$, $n_z$ resemble the discretization in vertical, horizontal and depth-direction, respectively. Equivalently, we may represent $\data_c$ by considering the real and imaginary parts as two separate channels, resulting a 4D real data tensor $\data \in \mathbb{R}^{n_x \times n_y \times n_z \times 2}$.
 
Since the system is calibrated by amplitude normalization with respect to an ideal metallic reflector, a rectangular frequency signal response is ensured for the FMCW frequency dependance~\cite{ding2013thz}.
After the FFT in~\eqref{eq:thzEquation_fft}, the $z$-direction signal envelope is an ideal $\sinc$ function as continuous spatial signal amplitude, giving rise to the physical model given in~\eqref{eq:thzEquation} in the introduction. 

In~\eqref{eq:thzEquation}, the electric field amplitude $\hat{e}$ is the reflection coefficient for the material, which is dependent on the complex dielectric constant of the material and helps to identify and classify materials. The depth position $\mu$ is the position at which maximum reflection occurs, i.e., the position of the surface reflecting the THz energy. $\sigma$ is the width of the reflected pulse, which includes information on the dispersion characteristics of the material. The phase $\phi$ of the reflected wave depends on the ratio of real to imaginary parts of the dielectric properties of the material. Thus, the parameters $\param=(\hat{e},\sigma, \mu,\phi)$ contain important information about the geometry as well as the material of the imaged object, which is of interest in a wide variety of applications. 

\section{Related Work}
\label{sec:related}
Due to the revolutionary success (convolutional) neural networks have had on computer vision problems over the last decade, researchers have extended the fields of applications of neural networks significantly. A particularly interesting concept is to learn the solution of complex, possibly nonconvex, optimization problems. Different lines of research have considered directly learning the optimizer itself, e.g. modelled as a recurrent neural network~\cite{Andrychowicz2016LearningTL}, or rolling out optimization algorithms and learning the incremental steps, e.g. in the form of parameterized proximal operators in \cite{Kobler2017VariationalNC}. Further hybrid approaches include optimization problems in the networks' architecture, e.g. \cite{Amos2017OptNetDO}, or combining optimizers with networks that have been trained individually \cite{Chang17,Meinhardt17}. The recent work of Moeller et al.~\cite{energyDissipation19} trains a network to predict descent directions to a given energy in order to give provable convergence results on the learned optimizer. 

Objectives similar to the one arising in the training of our model-based AEs are considered, for instance, for solving inverse problems with deep image priors~\cite{Ulyanov2018DeepIP} or deep decoders~\cite{Heckel2019DeepDC}. These works, however, consider the input to the networks being fixed random noise and have to solve an optimization problem for the networks weights for each inverse problems, such that they are regularization-by-parametrization approaches rather than learned optimizers. 

The most related prior work is the 3D face reconstruction network from Tewari\etal\cite{tewari17MoFA}. They aimed at finding a semantic code vector from a given facial image such that feeding this code vector into a rending engine yields an image similar to the input image itself. While this problem had been addressed using optimization algorithms a long time ago \cite{Blanz99} (also known under the name of analysis-by-synthesis approaches), the approach by Tewari\etal\cite{tewari17MoFA} replaced the optimizer with a neural network and kept the original cost function to train the network in an unsupervised way. The resulting structure resembles an AE in which the decoder fixed to the forward model and was therefore coined model-based AE. As we will discuss in the next section, the idea of model-based AEs generalizes far beyond 3D face reconstruction and can be used to boost the THz parameter identification problem significantly.

Finally, a recent work has exploited deep learning techniques in Terahertz imaging in~\cite{long2019terahertz}, but the considered application of super-resolving the THz amplitude image by training a convolutional neural network on synthetically blurred images is not directly related to our proposed approach. 

\section{A Model-Based Autoencoder for THz Image Reconstruction}
\label{sec:ourArchi}
Let us denote the THz input data by $\data \in \mathbb{R}^{n_x \times n_y \times n_z \times 2}$, and consider our four unknown parameters $(\hat{e},\sigma,\mu,\phi)$ to be $\mathbb{R}^{n_x \times n_y}$ matrices, allowing each parameter to change at each pixel. Under slight abuse of notation we can interpret all operations in~\eqref{eq:thzEquation} to be pointwise and again identify complex values with two real values in order to have $f_{\hat{e},\sigma,\mu,\omega,\phi}(z_{grid}) \in \mathbb{R}^{n_x \times n_y \times n_z \times 2}$, where $z_{grid} = (z_i)_{i \in \{1, \hdots, n_z\}}$ denotes the depth sampling grid. Concatenating all four matrix valued parameters into a single parameter tensor $\paramt \in \mathbb{R}^{n_y \times n_x \times 4}$, our goal can be formalized as finding $\paramt$ such that $f_{\paramt}(z_{grid})\approx \data$.

\begin{figure}[tb]
    \centering
    \includegraphics[width=1\textwidth]{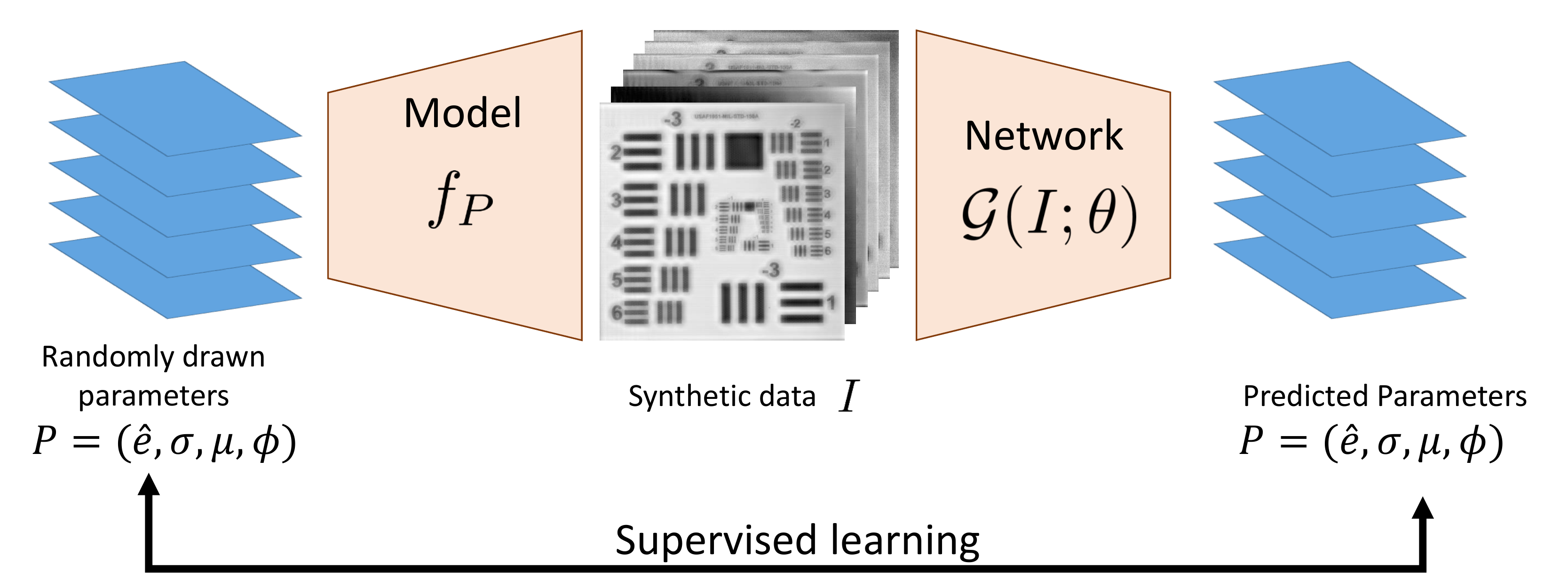}
    \caption{Classical supervised learning strategy with simulated data: The forward model $f_P$ (e.g. from \eqref{eq:thzEquation}) is used to simulate data $\data$, which can subsequently be fed into a network to be trained to reproduce the simulation parameters in a supervised way.}
    \label{fig:classicalArchi}
\end{figure}

A classical supervised machine learning approach to problems with known forward operator is illustrated in Figure \ref{fig:classicalArchi} for the example of THz image reconstruction: The explicit forward model $f$ is used to simulate a large set of images $\data$ from known parameters $P$ which can subsequently be used as training data for predicting $P$ via a neural network $\net(\data;\theta)$ depending on weights $\theta$. Such supervised approaches with simulated training data are frequently used in other image reconstruction areas, e.g. super resolution \cite{dong2014learning,kim2016accurate}, or image deblurring \cite{nah2017deep,schuler2016learning}. The accuracy of networks trained on simulated data, however, crucially relies on precise knowledge of the forward model and the simulated noise. Slight deviations thereof can significantly degrade a network performance as demonstrated in \cite{DarmstadtDenoising}, where deep denoising networks trained on Gaussian noise were outperformed by BM3D when applied to realistic sensor noise.

\begin{figure}[thb]
    \centering
    \includegraphics[width=1\textwidth]{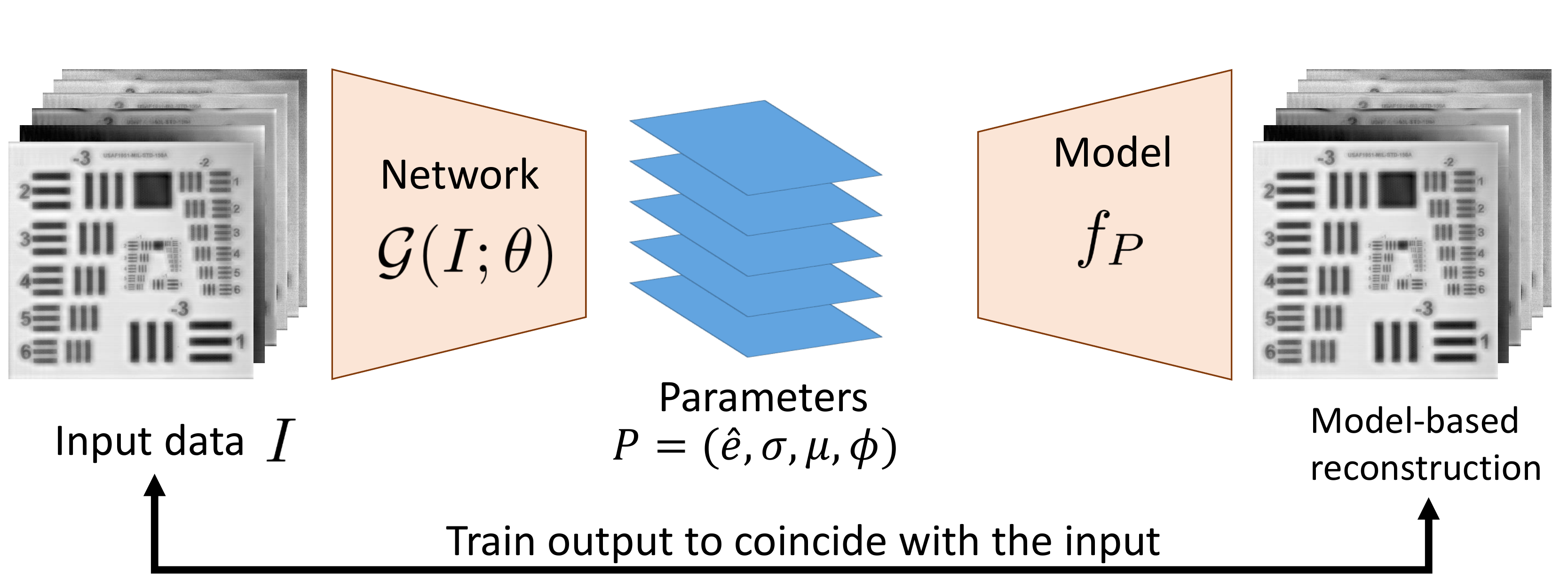}
    \caption{A model-based AE for THz image reconstruction: The input data $\data$ is fed into a network $\net$ whose parameters $\theta$ are trained in such a way that feeding the network's prediction $\net(\data;\theta)$ into a model function $f$ again reproduces the input data $\data$. Such an architecture resembles an AE with a learnable encoder and a model-based decoder and allows an unsupervised training on real data.}
    \label{fig:netArchi}
\end{figure}

Instead of pursuing the supervised learning approach described above, we replace $\param=(\hat{e},\sigma, \mu, \phi)$ in the optimization approach \eqref{eq:thzOptimization} by a suitable network $\net(\data;\theta)$ that depends on the raw input data $\data$ and learnable parameters $\theta$, that can be trained in an \textit{unsupervised} way \textit{on real data}. Assuming we have multiple examples $\data^k$ of THz data, and choosing the loss function in \eqref{eq:thzOptimization} as an $\ell^2$-squared loss, gives rise to the unsupervised training problem
\begin{equation}
\label{eq:training}
\min_\theta \sum_{\text{training examples }k} \|f_{\net(\data^k;\theta)}(z_{grid}) - \data^k\|_F^2.
\end{equation}
As we have illustrated in Figure \ref{fig:netArchi}, this training resembles an AE architecture: The input to the network is data $\data^k$ which gets mapped to parameters $P$ that -- when fed into the model function $f$ -- ought to reproduce $\data^k$ again.

Opposed to the straight forward supervised learning approach, the proposed approach \eqref{eq:training} has two significant advantages
\begin{itemize}
\item It allows us to train the network in an \textit{unsupervised} way, i.e., on real data, and therefore learn to deal with measurement-specific distortions.
\item The cost function in \eqref{eq:training} implicitly handles the scaling of different parameters, and therefore circumvents the problem of defining meaningful cost functions on the parameter space: Simple parameter discrepancies such as $\|P_1  - P_2\|_2^2$ for two different parameters sets $P_1$ and $P_2$ largely depend on the scaling of the individual parameters and might even be meaningless, e.g. for cyclic parameters such as the phase offset $\phi$. 
\end{itemize}

\section{Encoder Network Architecture and Training}
\label{sec:implementation}

\subsection{Data Preprocessing}
\label{sec:preproc}

\begin{figure}[tb]
    \centering
    \includegraphics[width=0.5\textwidth]{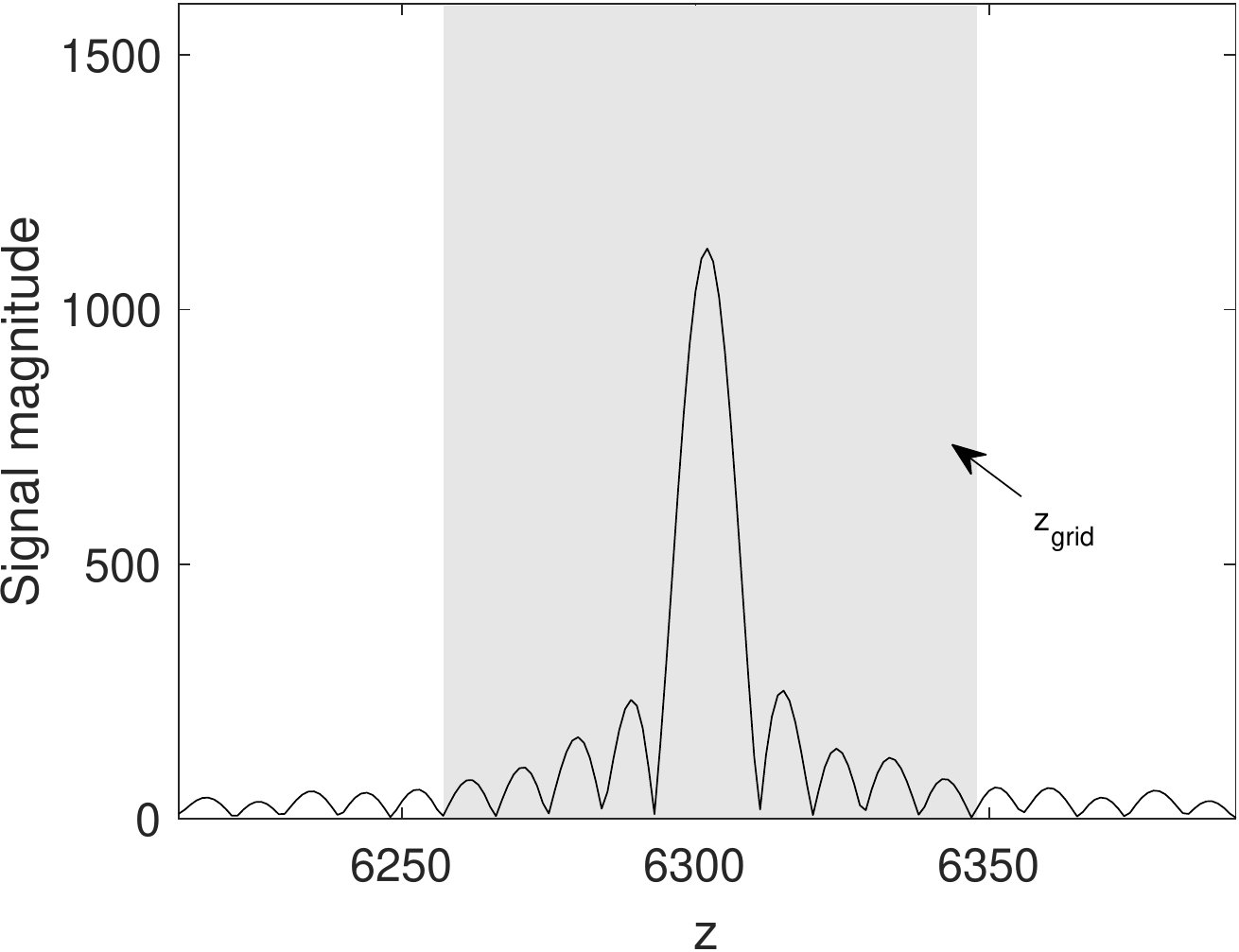}
    \caption{Magnitude of a sample point of measured THz signal. The main lobe and major side-lobes are included in the grid window, which is colored in gray.}
    \label{fig:exemplarySignal}
\end{figure}

As illustrated in the plot of the magnitude of an exemplary measured THz signal shown in Figure \ref{fig:exemplarySignal}, the THz energy is mainly focused in the main lobe and first side-lobes of the $\sinc$ function. Because the physical model remains valid in close proximity of the main lobe only, we preprocess the data to reduce the impressively large range of $12600$ measurements per pixel. We, therefore, crop out 91 measurements per pixel centered around the main lobe, whose position is related to the object distance and to the parameter $\mu$.  Details of the cropping window are described in~\cite{wong2019computational}. We represent the THz data in a 4D real tensor $\data \in \mathbb{R}^{n_x \times n_y \times n_z \times 2}$, where $n_x = n_y = 446$, and $n_z$ is the size of the cropping window, i.e. $91$ in our case. 

\subsection{Encoder Architecture and Training}
\begin{figure}[htb]
    \centering
    \includegraphics[width = 1\textwidth]{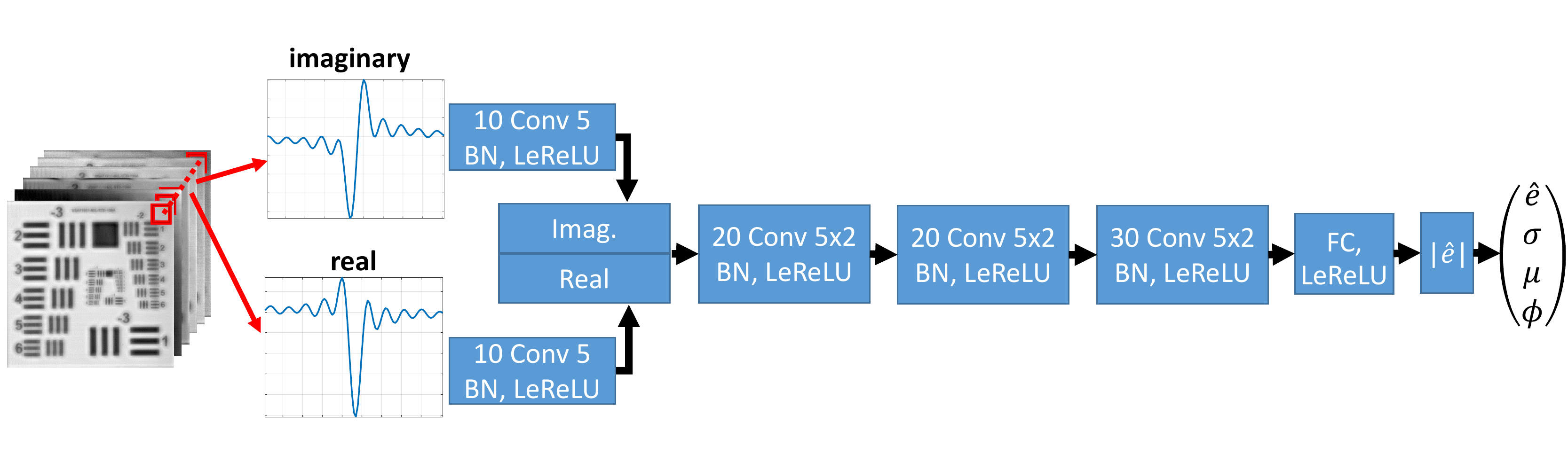}
    \caption{Architecture of encoding network $\net(\data;\theta)$ that predicts the parameters: At each pixel the real and imaginary part is extracted, convolved, concatenated and processed via three convolutional and 1 fully connected layer. To obtain physically meaningful (non-negative) amplitudes, we apply an absolute value function to the first component. }
    \label{fig:encoder}
\end{figure}

For the encoder network $\net(\data;\theta)$ we pick a spatially decoupled architecture using $1\times 1$ convolutions on $\data$ only, leading to a signal-by-signal reconstruction mechanism that allows a high level of parallelism and therefore maximizes the reconstruction speed on a GPU. The specific architecture (illustrated in Figure~\ref{fig:encoder}) applies a first set of convolutional filters on the real and imaginary part separately, before concatenating the activations, and applying three further convolutional filters on the concatenated structure. We apply batch-normalization (BN)~\cite{ioffe2015batch} after each convolution and use leaky rectified linear units (LeReLU)~\cite{glorot2011deep} as activations. Finally, a fully connected layer reduces the dimension to the desired size of four output parameters per pixel. To ensure that the amplitude is physically meaningful, i.e., non-negative, we apply an absolute value function on the first component. Interestingly, this choice compared favorably to a plain rectified linear unit when the network is trained. 

\begin{figure}[htb]
    \centering
    \includegraphics[width=0.5\textwidth]{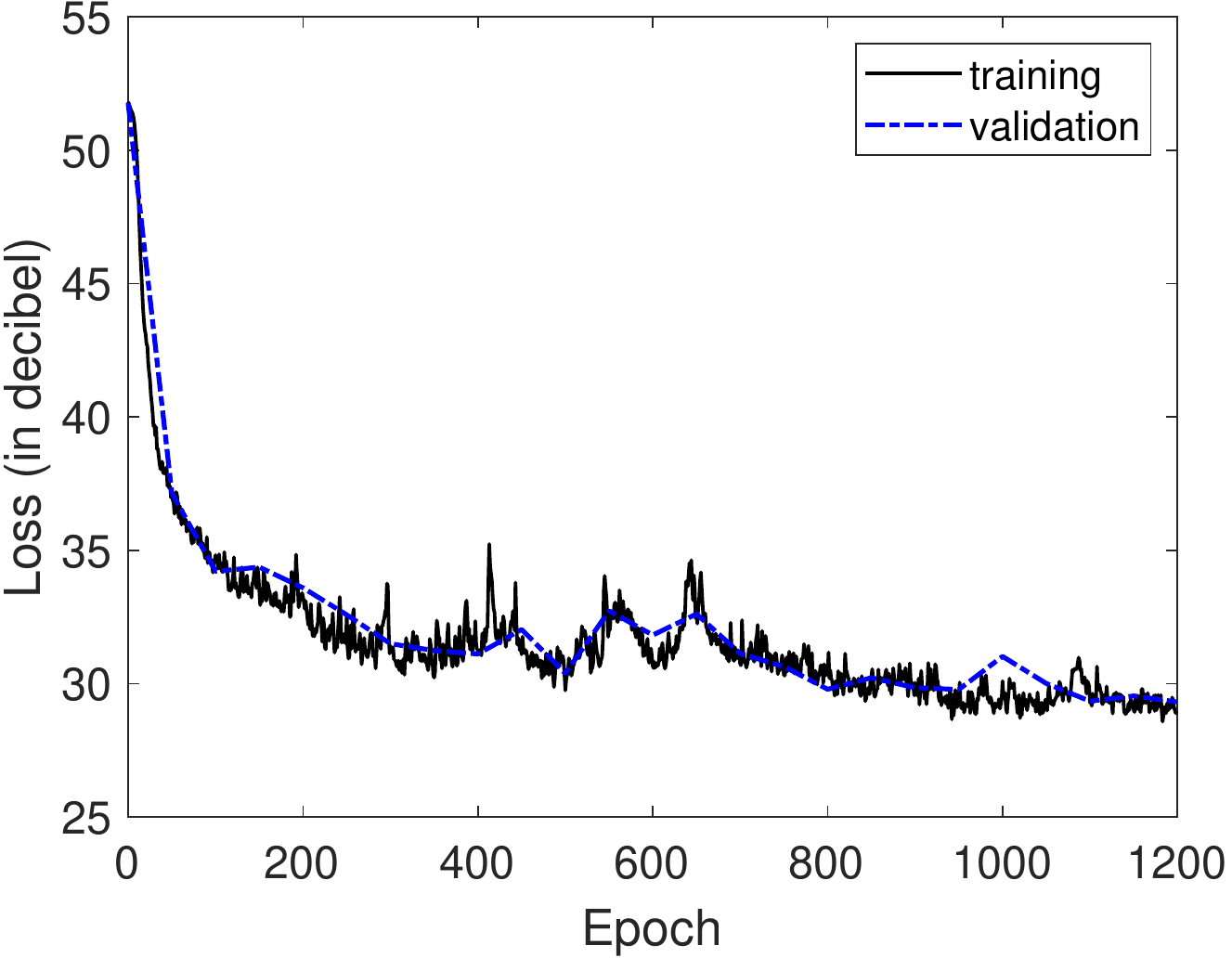}
    \caption{
    \protect The average losses of the training and validation sets over 1200 epochs on a decibel scale illustrate that there is almost no generalization gap between training and validation. 
    }
    \label{fig:training}
\end{figure}

We train our model optimizing \eqref{eq:training} using the Adam optimizer~\cite{kingma2014adam} on $80\%$ of the $446 \times 446$ pixels from a real (measured) THz image for 1200 epochs. The remaining $20\%$ of the pixels serve as a validation set. The batch size is set to $4096$. The initial learning rate is set to $0.005$, and is reduced by a factor of 0.99 every 20 epochs. Figure~\ref{fig:training} illustrates the decay of the training and validation losses over 1200 epochs. As we can see, the validation loss nicely resembles the training loss with almost no generalization gap. 

\section{Numerical Experiments}
\label{sec:experiments}

\begin{figure}[thb]
    \centering
    \subfloat[\label{fig:thz_object_metalpcb}]{\includegraphics[width=0.25\textwidth]{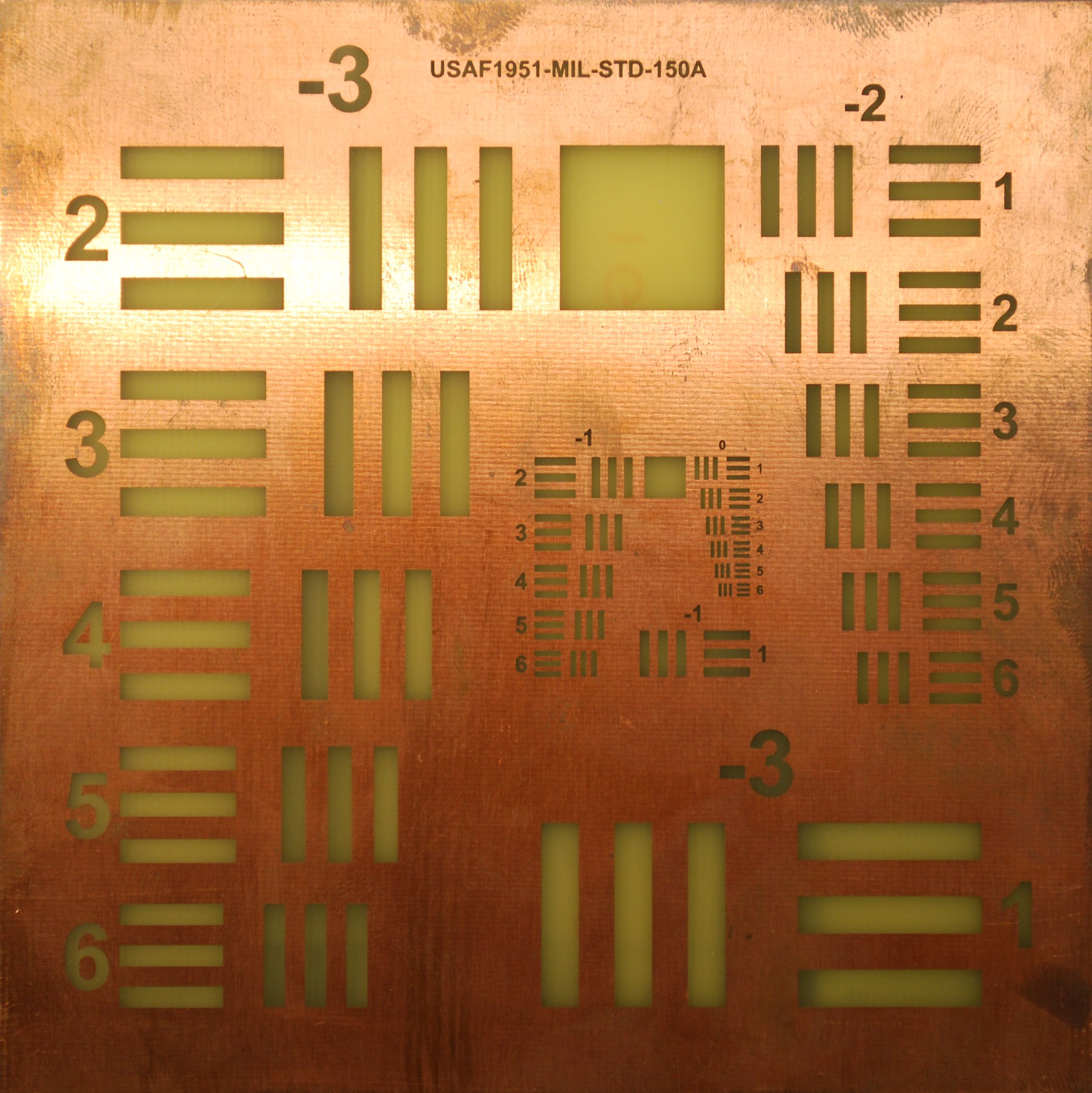}}
    \hspace{0.1\textwidth}
    \subfloat[\label{fig:thz_object_stepchart}]{\includegraphics[width=0.25\textwidth]{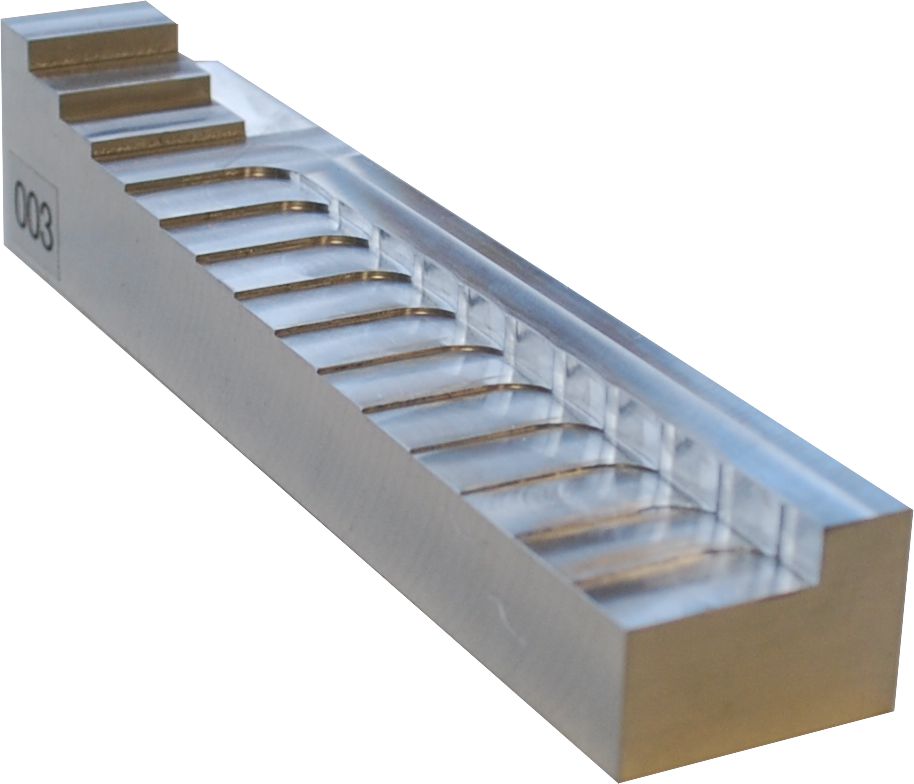}}
    \caption{Objects of evaluated datasets \protect\subref{fig:thz_object_metalpcb} MetalPCB dataset
    \protect\subref{fig:thz_object_stepchart} StepChart dataset
    }
    \label{fig:thz_object}
\end{figure}
We evaluate the proposed model-based AE on two datasets, which are acquired using the setup described in Sec.~\ref{sec:THz}, namely the \emph{MetalPCB} dataset and the \emph{StepChart} dataset. 
The MetalPCB dataset is measured by a nearly planar copper target etched on a circuit board (Figure~\ref{fig:thz_object_metalpcb}), 
which includes \emph{metal} and \emph{PCB} material regions, 
in the standard size scale of USAF target \emph{MIL-STD-150A}~\cite{standard1959photographic}. 
After the preprocessing described in Sec.~\ref{sec:preproc}, the MetalPCB dataset has $446 \times 446 \times 91$ sample points.
The StepChart dataset is based on an aluminum object (Figure~\ref{fig:thz_object_stepchart}) with sharp edges to evaluate the distance measurement accuracy using a 3D object.
The StepChart dataset has $113 \times 575 \times 91$ sample points after preprocessing.

In order to evaluate the optimization quality on different materials and structures, 
MetalPCB dataset is evaluated in regions: 
\emph{PCB region} is a local region that contains PCB material only,
\emph{Metal region} is a local region that contains copper material only, 
and \emph{All region} is the entire image area.
Similarly, the StepChart dataset is evaluated by 3 regions: 
\emph{Edge region} is the region that contains physical edges, 
\emph{Steps region} is the center planar region of each steps, 
and \emph{All region} is the entire image area. This segmentation is done, because the THz measurements of the highly specular aluminum target results in strong multi-path interference artifacts at the edges that should be investigated separately.

The proposed model-based AE is trained on the MetalPCB dataset only, while the parameter inference is made for both the MetalPCB and StepChart datasets. 
This cross-referencing between two datasets can verify whether the proposed AE method is modelling the physical behavior of the system without overfitting to a specific dataset or recorded material. 

To compare with the classical optimization methods, the parameters are estimated using the Trust-Region Algorithm (TRA)~\cite{coleman1996interior}, which is implemented in MATLAB{\tm}. The TRA optimization requires a proper definition of the parameter ranges. Furthermore, it is very sensitive with respect to the initial parameter set. We, therefore, carefully select the initial parameters by sequentially estimating them from the source data (see~\cite{wong2019computational} for more details). Still, the optimization may result in a parameter set with significant loss values; see Sec.~\ref{sec:experiments.quality}.

The trained encoder network is independent of any initialization scheme as it tries to directly predict optimal parameters from the input data. While the network alone gives remarkably good results with significantly lower runtimes than the optimization method, there is no guarantee that the network's predictions are critical points of the energy to be minimized. This motivates the use of the encoder network as an initialization scheme to the TRA, specifically because the TRA guarantees the monotonic decrease of objective function such that using the TRA on top of the network can only improve the results. We abbreviate this approach to \emph{AE+TRA} for the rest of this paper. 

To fairly compare all three approaches, the optimization time of TRA and the inference time of the AE are both recorded by an Intel\tm i7-8700K CPU computation, while the AE is trained on a NVIDIA\tm GTX 1080 GPU. 
The PyTorch source code is available at \url{https://github.com/tak-wong/THz-AutoEncoder}.

\subsection{Loss and timing}
\label{sec:experiments.loss}

\begin{table}[htb]
\renewcommand{\arraystretch}{1.1}
\centering
\caption{Loss and timing enhancement based on the proposed model-based AE}
\label{tab:loss}
\begin{tabular}{|l|l|r|r|r|}
\hline
Dataset (Region) & Measurement & TRA & AE & AE+TRA \\ \hline
MetalPCB (All) & Average Loss & 693.9 & 886.3 & \textbf{442.2} \\
MetalPCB (PCB) & Average Loss & \textbf{589.0} & 872.6 & \textbf{589.0} \\
MetalPCB (Metal) & Average Loss & 519.6 & 446.1 & \textbf{115.7} \\
StepChart (All) & Average Loss & 3815.1 & 5148.3 & \textbf{3675.3} \\ 
StepChart (Edges) & Average Loss & 4860.4 & 6309.1 & \textbf{2015.7} \\
StepChart (Steps) & Average Loss & 1152.5 & 2015.7 & \textbf{1150.3} \\ \hline
MetalPCB & Training time (sec.) & \textbf{none}  & {9312.8} & {9312.8} \\
MetalPCB & Run time (sec.) & 10391.2 & $^{\dagger}$\textbf{73.5} & $^*$4854.7 \\
StepChart & Run time (sec.) & 3463.9 & $^{\dagger}$\textbf{22.8} & $^*$1712.4 \\ \hline
\multicolumn{5}{l}{\footnotesize{$^{\dagger}$ Inference time}} \\
\multicolumn{5}{l}{\footnotesize{$^*$ Run time is the sum of AE inference and TRA optimization time}} \\
\end{tabular}
\end{table}

In Table~\ref{tab:loss}, the average loss in \eqref{eq:training} and the timing are shown for the Trust-Region Algorithm (TRA), the Autoencoder (AE) and the joint AE+TRA approaches, respectively. We can see that the proposed encoder network achieves a lower average loss than the TRA method in the metal region of the MetalPCB dataset, it yields higher average losses than the TRA on both datasets. 
It is encouraging to see that although the AE was trained on the MetalPCB dataset, the relative performance in comparison to the TRA does not decay too significantly when changing to an entirely unseen data set with a different material, with the AE loss being $21.7\%$ and $25.9\%$ higher than the TRA loss on the MetalPCB and StepChart data sets, respectively. If such a sacrifice in accuracy is acceptable, the speed-up in runtime is tremendous with the AE being over 140 times faster than the TRA (for both methods being evaluated on a CPU). Note that even the sum of training and inference time are smaller for the proposed AE than the runtime of the TRA on the MetalPCB dataset. 

Interestingly, the combined AE+TRA approach of initializing the TRA with the encoder network's prediction leads to better losses than the TRA alone in all regions. Additionally, the AE-initialized TRA converged more than 2 times faster due to the stopping criterion being reached earlier.

We note that the losses of all approaches are significantly higher for the StepCart data set than they are for the MetalPCB. 
This is because the aluminum StepChart object (Figure~\ref{fig:thz_object_stepchart}) has a more complex physical structure than the MetalPCB object, which results in a mixture of scattered THz pulses by multi-path interference effects in all object regions. Incorporating such effects in the reflection model of \eqref{eq:thzEquation} could therefore be an interesting aspect of future research for improving the explainability of the measured data with the physical model. 

\subsection{Quality Assessment of THz Images}
\label{sec:experiments.quality}

\begin{figure}[!h]
	\centering
	\subfloat[\label{fig:intensity_source}]{\includegraphics[width=0.49\textwidth]{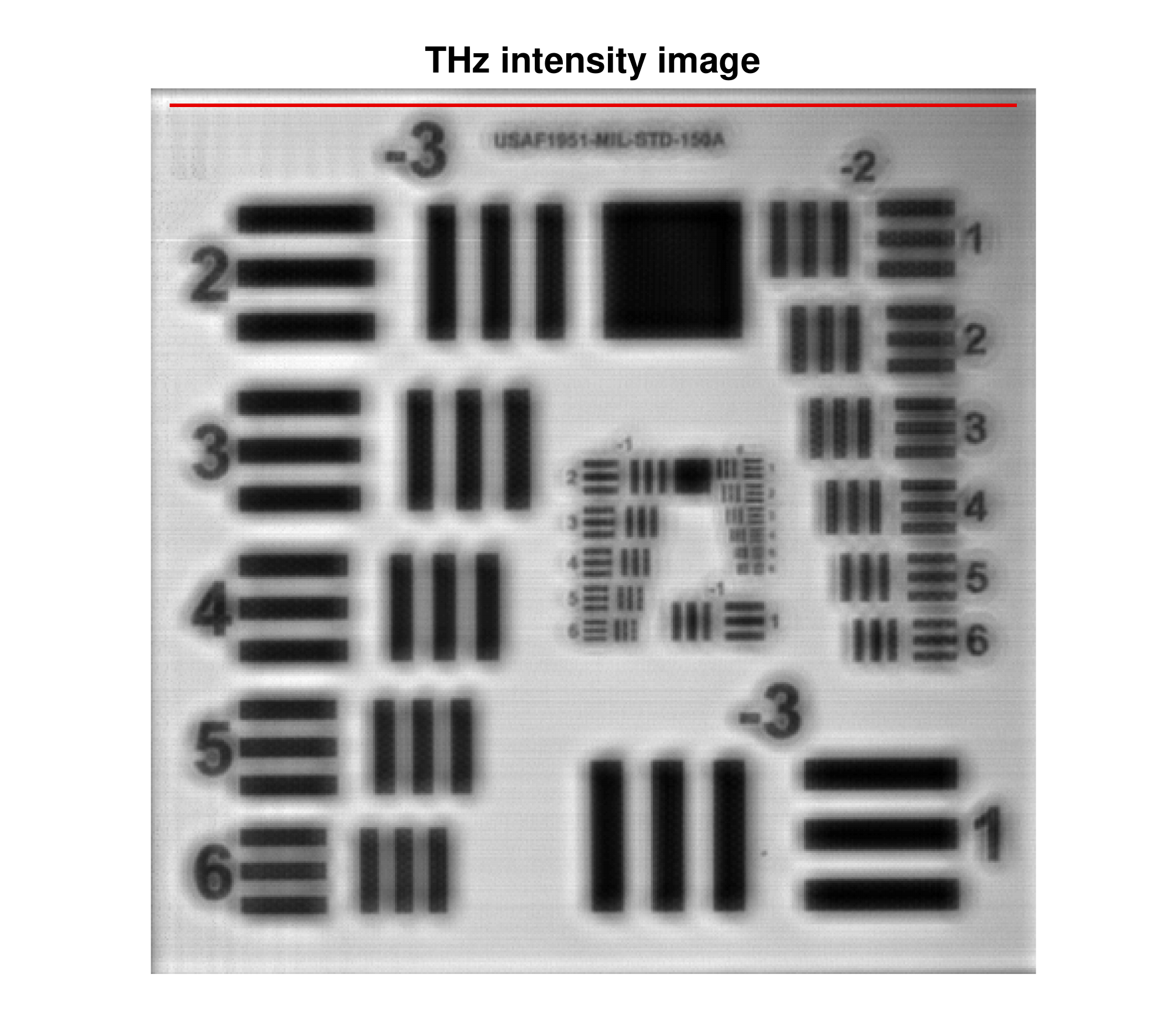}}
	\hfill
	\subfloat[\label{fig:intensity_hybrid}]{\includegraphics[width=0.49\textwidth]{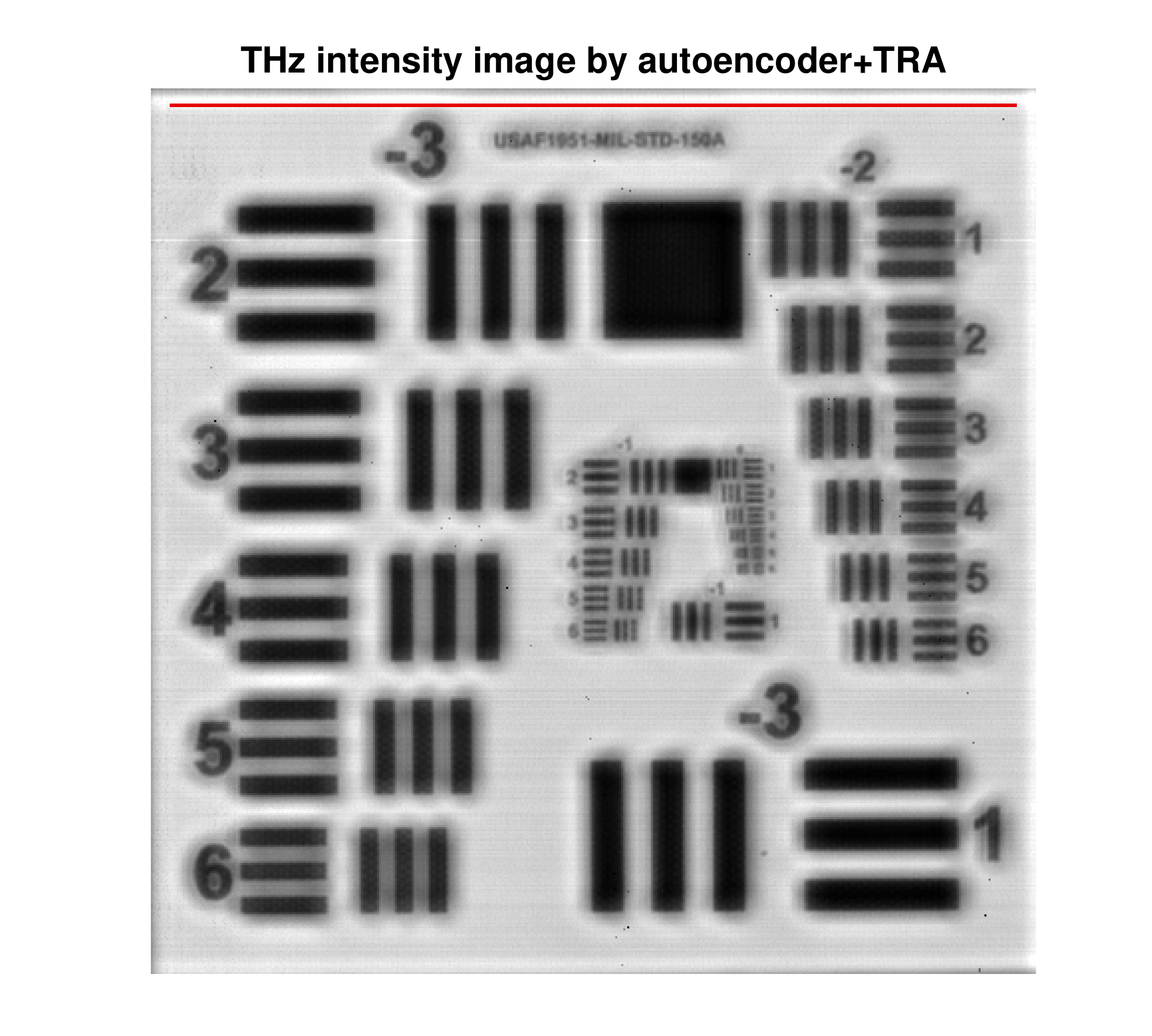}}
	\vfill
	\subfloat[\label{fig:line_intensity}]{\includegraphics[width=0.49\textwidth]{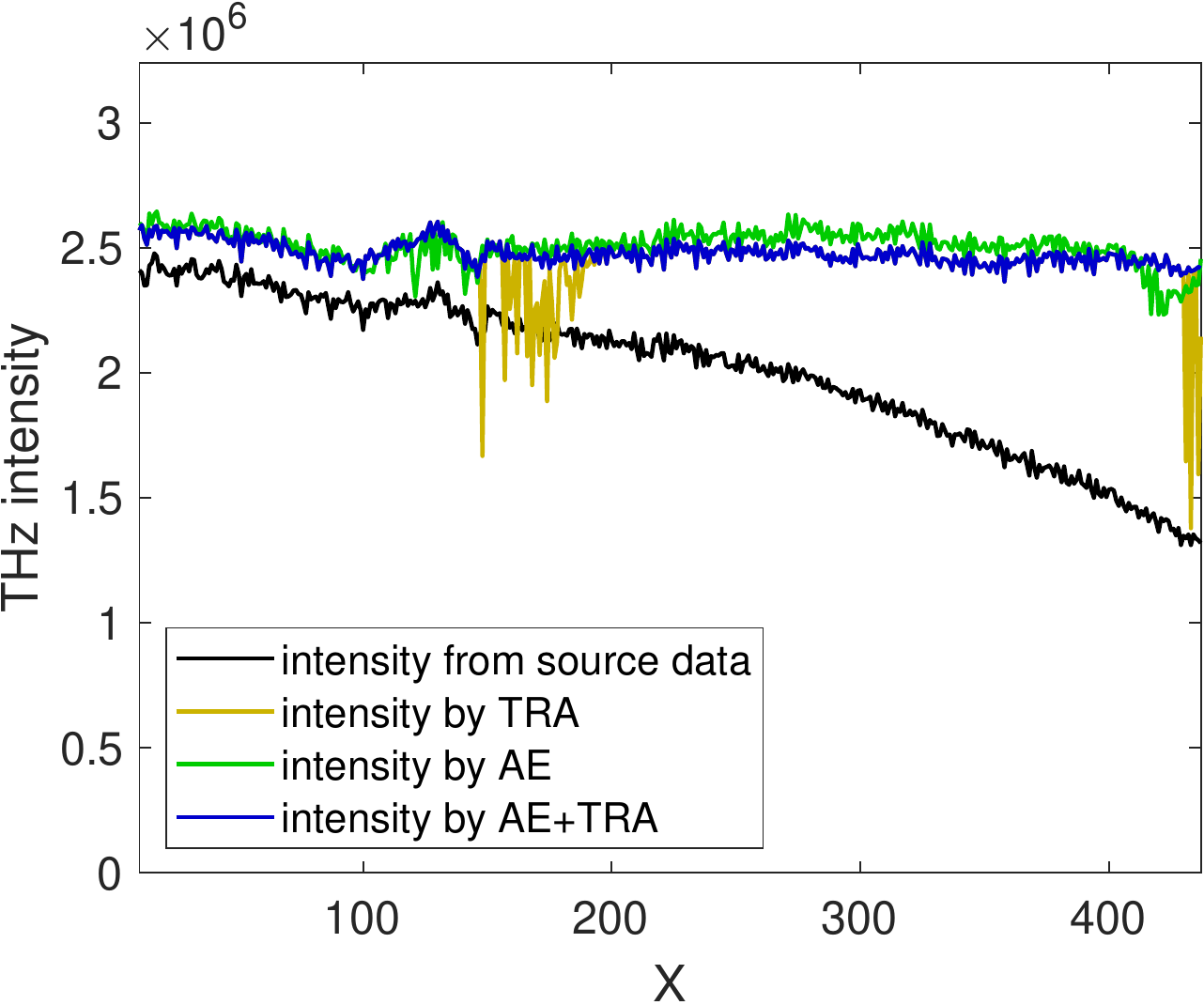}}
	\hfill
	\subfloat[\label{fig:line_loss}]{\includegraphics[width=0.49\textwidth]{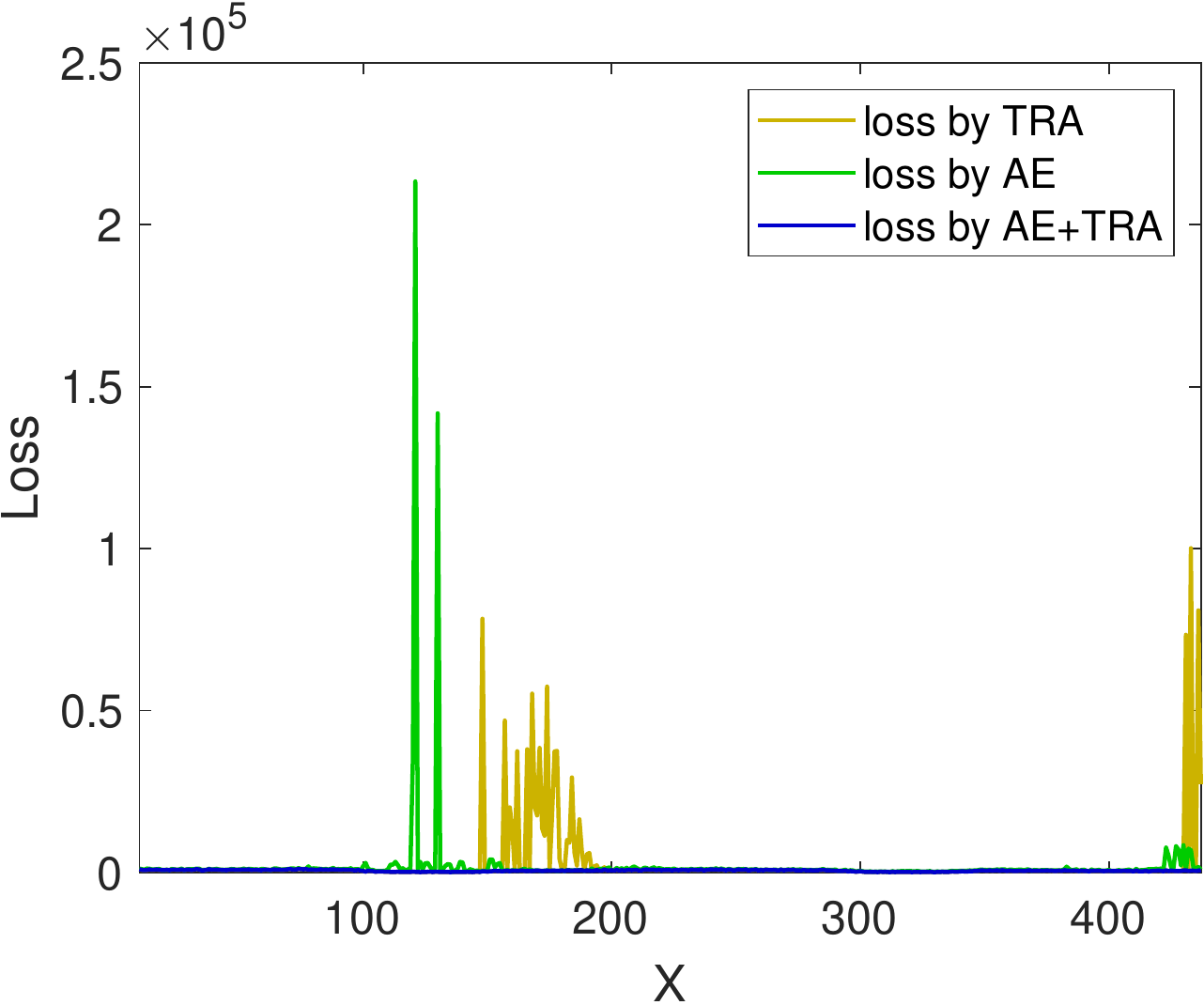}}
	\caption{Comparison of the THz intensity for the MetalPCB dataset: 
	\protect\subref{fig:intensity_source} intensity image extracted from the source data without any model-based processing (in red: the pixel line for plots \protect\subref{fig:line_intensity} and \protect\subref{fig:line_loss});
	\protect\subref{fig:intensity_hybrid} image extracted by the proposed AE+TRA approach (in red: the pixel line for plots \protect\subref{fig:line_intensity} and \protect\subref{fig:line_loss});
	\protect\subref{fig:line_intensity} plot of the intensity extracted along the horizontal line in the copper region;
    \protect\subref{fig:line_loss} plot of the per-pixel loss by TRA, AE, and AE+TRA approaches along the horizontal line in the copper region.}
	\label{fig:intensity_comparison}
\end{figure}

In THz imaging, the \emph{intensity} image $I$ that is equal to the squared amplitude, i.e. $I=\hat{e}^2$ is the most important criteria for quality assessment. 
Note that the intensity could be inferred directly from the data by considering that \eqref{eq:thzEquation} yields 
\begin{equation}
\label{eq:thzEquation_intensity}
\fun_{\hat{e},\sigma,\mu,\phi}(\mu) \cdot \fun^*_{\hat{e},\sigma,\mu,\phi}(\mu) = \hat{e}^2 \cdot {\sinc}^2(0) = \hat{e}^2 = I
\end{equation}
where $\fun^*$ is the complex conjugate of $\fun$.
As we illustrate in Figure \ref{fig:intensity_comparison}, the model-based approach is not only capable of extracting all relevant parameters, \ie, $\hat{e}$, $\mu$, $\sigma$ and $\phi$, but, compared to values directly extracted from the source data, the resulting intensity $I$ is more homogeneous in homogeneous material regions. The homogeneity of the directly extracted intensity results from the very low depth of field of THz imaging systems in general, combined with the slight non-planarity of the MetalPCB target. As depicted in Figure~\ref{fig:line_intensity}, the intensity variations along the selected line in the homogeneous copper region are reduced using the three model-based methods, \ie TRA, AE, and AE+TRA. However, due to the crucial selection of the initial parameters (see discussion at the beginning of Sec.~\ref{sec:experiments}), the TRA optimization results exhibit significant amplitude fluctuations and loss values (Figure \ref{fig:line_loss}) in the two horizontal sub-regions  $x\in[150,200]$ and $x>430$. The proposed AE and AE+TRA methods, however, deliver superior results with respect to the main quality measure applied in THz imaging, \ie to the intensity homogeneity and the loss in model fitting. Still, the AE approach shows very few extreme loss values, while the AE+TRA method's loss values are consistently low along the selected line in the homogeneous copper region.

\section{Conclusions and Future Work}
\label{sec:conclusions}
In this paper, we propose a model-based autoencoder for THz image reconstruction.
Comparing to a classical Trust-Region optimizer, the proposed autoencoder gets within $25\%$ margin to the objective value of the optimizer, while being more than 140 times faster. Using the network's prediction as an initialization to a gradient-based optimization scheme improves the result over a plain optimization scheme in terms of objective values while still being two times faster. We believe that these are very promising results for training optimizers/initialization schemes for parameter identification problems in general by exploiting the idea of model-based autoencoders for unsupervised learning. 

 Future research will include exploiting spatial information during the reconstruction as well as considering joint parameter identification and reconstruction problems such as denoising, sharpening, and super-resolving parameter images such as the amplitude images shown in Figure \ref{fig:intensity_hybrid}.
 

\section*{Acknowledgement}

This is a pre-print of a conference proceeding article published in German Conference on Pattern Recognition. The final authenticated version is available online at: \href{https://doi.org/10.1007/978-3-030-33676-9\_7}{https://doi.org/10.1007/978-3-030-33676-9\_7}


\bibliographystyle{unsrt}
\bibliography{bibliography}

\begin{thebibliography}{10}

\bibitem{chan2007imaging}
Wai~Lam Chan, Jason Deibel, and Daniel~M Mittleman.
\newblock Imaging with terahertz radiation.
\newblock {\em Reports on progress in physics}, 70(8):1325, 2007.

\bibitem{Jansen2010}
Christian Jansen, Steffen Wietzke, Ole Peters, Maik Scheller, Nico Vieweg,
  Mohammed Salhi, Norman Krumbholz, Christian J\"{o}rdens, Thomas Hochrein, and
  Martin Koch.
\newblock Terahertz imaging: applications and perspectives.
\newblock {\em Appl. Opt.}, 49(19):E48--E57, 2010.

\bibitem{siegel2002terahertz}
Peter~H Siegel.
\newblock Terahertz technology.
\newblock {\em IEEE Transactions on microwave theory and techniques},
  50(3):910--928, 2002.

\bibitem{wong2019computational}
Tak~Ming Wong, Matthias Kahl, Peter Haring~Bol{\'i}var, and Andreas Kolb.
\newblock Computational image enhancement for frequency modulated continuous
  wave (fmcw) thz image.
\newblock {\em Journal of Infrared, Millimeter, and Terahertz Waves},
  40(7):775--800, 2019.

\bibitem{cooper2011thz}
Ken~B Cooper, Robert~J Dengler, Nuria Llombart, Bertrand Thomas, Goutam
  Chattopadhyay, and Peter~H Siegel.
\newblock Thz imaging radar for standoff personnel screening.
\newblock {\em IEEE Transactions on Terahertz Science and Technology},
  1(1):169--182, 2011.

\bibitem{hu1995imaging}
Binbin~B Hu and Martin~C Nuss.
\newblock Imaging with terahertz waves.
\newblock {\em Optics letters}, 20(16):1716--1718, 1995.

\bibitem{mcclatchey2001time}
K~McClatchey, MT~Reiten, and RA~Cheville.
\newblock Time resolved synthetic aperture terahertz impulse imaging.
\newblock {\em Applied physics letters}, 79(27):4485--4487, 2001.

\bibitem{ding2013thz}
Jinshan Ding, Matthias Kahl, Otmar Loffeld, and Peter Haring~Bol{\'\i}var.
\newblock Thz 3-d image formation using sar techniques: simulation, processing
  and experimental results.
\newblock {\em IEEE Transactions on Terahertz Science and Technology},
  3(5):606--616, 2013.

\bibitem{kahl2012}
M.~Kahl, A.~Keil, J.~Peuser, T.~L\"offler, M.~P\"atzold, A.~Kolb, T.~Sprenger,
  B.~Hils, and P.~Haring~Bol{\'i}var.
\newblock Stand-off real-time synthetic imaging at mm-wave frequencies.
\newblock In {\em Passive and Active Millimeter-Wave Imaging XV}, volume 8362,
  page 836208, 2012.

\bibitem{munson1989signal}
David~C Munson and Robert~L Visentin.
\newblock A signal processing view of strip-mapping synthetic aperture radar.
\newblock {\em IEEE Transactions on Acoustics, Speech, and Signal Processing},
  37(12):2131--2147, 1989.

\bibitem{skolnik1970radar}
Merrill~Ivan Skolnik.
\newblock Radar handbook.
\newblock 1970.

\bibitem{Andrychowicz2016LearningTL}
Marcin Andrychowicz, Misha Denil, Sergio~Gomez Colmenarejo, Matthew~W. Hoffman,
  David Pfau, Tom Schaul, and Nando de~Freitas.
\newblock Learning to learn by gradient descent by gradient descent.
\newblock In {\em Proc. Int. Conf. on Neural Information Processing Systems
  (NIPS)}, 2016.

\bibitem{Kobler2017VariationalNC}
Erich Kobler, Teresa Klatzer, Kerstin Hammernik, and Thomas Pock.
\newblock Variational networks: Connecting variational methods and deep
  learning.
\newblock In {\em Proc. German Conf. Pattern Recognition (GCPR)}, 2017.

\bibitem{Amos2017OptNetDO}
Brandon Amos and J.~Zico Kolter.
\newblock Optnet: Differentiable optimization as a layer in neural networks.
\newblock In {\em Proc. Int. Conf. on Machine Learning}, 2017.

\bibitem{Chang17}
J-H. Chang, C-L. Li, B.~Poczos, B.V.K.~Vijaya Kumar, and A.C. Sankaranarayanan.
\newblock One network to solve them all --- solving linear inverse problems
  using deep projection models.
\newblock In {\em Proc. IEEE Int. Conf. on Computer Vision}, 2017.

\bibitem{Meinhardt17}
T.~Meinhardt, M.~Moeller, C.~Hazirbas, and D.~Cremers.
\newblock Learning proximal operators: Using denoising networks for
  regularizing inverse imaging problems.
\newblock In {\em Proc. IEEE Int. Conf. on Computer Vision}, 2017.

\bibitem{energyDissipation19}
M.~Moeller, T.~M\"ollenhoff, and D.~Cremers.
\newblock Controlling neural networks via energy dissipation, 2019.
\newblock Online at \url{https://arxiv.org/abs/1904.03081}.

\bibitem{Ulyanov2018DeepIP}
D.~Ulyanov, A.~Vedaldi, and V.S. Lempitsky.
\newblock Deep image prior.
\newblock In {\em Proc. IEEE Conf. Computer Vision and Pattern Recognition},
  2018.

\bibitem{Heckel2019DeepDC}
R.~Heckel and P.~Hand.
\newblock Deep decoder: Concise image representations from untrained
  non-convolutional networks.
\newblock In {\em Int. Conf. on Learning Representations}, 2019.

\bibitem{tewari17MoFA}
Ayush Tewari, Michael Zoll{\"o}fer, Hyeongwoo Kim, Pablo Garrido, Florian
  Bernard, Patrick Perez, and Theobalt Christian.
\newblock {MoFA: Model-based Deep Convolutional Face Autoencoder for
  Unsupervised Monocular Reconstruction}.
\newblock In {\em Proc. IEEE Int. Conf. on Computer Vision}, 2017.

\bibitem{Blanz99}
Volker Blanz and Thomas Vetter.
\newblock A morphable model for the synthesis of 3d faces.
\newblock In {\em Proc. {SIGGRAPH}}, pages 187--194, New York, NY, USA, 1999.
  ACM Press/Addison-Wesley Publishing Co.

\bibitem{long2019terahertz}
Zhenyu Long, Tianyi Wang, ChengWu You, Zhengang Yang, Kejia Wang, and Jinsong
  Liu.
\newblock Terahertz image super-resolution based on a deep convolutional neural
  network.
\newblock {\em Applied Optics}, 58(10):2731--2735, 2019.

\bibitem{dong2014learning}
Chao Dong, Chen~Change Loy, Kaiming He, and Xiaoou Tang.
\newblock Learning a deep convolutional network for image super-resolution.
\newblock In {\em Proc. Europ. Conf. Computer Vision}, pages 184--199.
  Springer, 2014.

\bibitem{kim2016accurate}
Jiwon Kim, Jung Kwon~Lee, and Kyoung Mu~Lee.
\newblock Accurate image super-resolution using very deep convolutional
  networks.
\newblock In {\em Proc. IEEE Conf. Computer Vision and Pattern Recognition},
  pages 1646--1654, 2016.

\bibitem{nah2017deep}
Seungjun Nah, Tae Hyun~Kim, and Kyoung Mu~Lee.
\newblock Deep multi-scale convolutional neural network for dynamic scene
  deblurring.
\newblock In {\em Proc. IEEE Conf. Computer Vision and Pattern Recognition},
  pages 3883--3891, 2017.

\bibitem{schuler2016learning}
Christian~J Schuler, Michael Hirsch, Stefan Harmeling, and Bernhard
  Sch{\"o}lkopf.
\newblock Learning to deblur.
\newblock {\em {IEEE} Trans. Pattern Analysis and Machine Intelligence (PAMI)},
  38(7):1439--1451, 2016.

\bibitem{DarmstadtDenoising}
T.~Pl\"otz and S.~Roth.
\newblock Benchmarking denoising algorithms with real photographs.
\newblock In {\em Proc. IEEE Conf. Computer Vision and Pattern Recognition},
  2017.

\bibitem{ioffe2015batch}
Sergey Ioffe and Christian Szegedy.
\newblock Batch normalization: Accelerating deep network training by reducing
  internal covariate shift.
\newblock {\em Proc. Int. Conf. on Machine Learning}, 2015.

\bibitem{glorot2011deep}
Xavier Glorot, Antoine Bordes, and Yoshua Bengio.
\newblock Deep sparse rectifier neural networks.
\newblock In {\em Proceedings of the fourteenth international conference on
  artificial intelligence and statistics}, pages 315--323, 2011.

\bibitem{kingma2014adam}
Diederik~P Kingma and Jimmy Ba.
\newblock Adam: A method for stochastic optimization.
\newblock {\em arXiv preprint arXiv:1412.6980}, 2014.

\bibitem{standard1959photographic}
Military Standard.
\newblock Photographic lenses, 1959.

\bibitem{coleman1996interior}
Thomas~F Coleman and Yuying Li.
\newblock An interior trust region approach for nonlinear minimization subject
  to bounds.
\newblock {\em SIAM Journal on optimization}, 6(2):418--445, 1996.

\end{thebibliography}

\end{document}